\documentclass[sigconf]{acmart}

\usepackage{hyperref} 
\usepackage{natbib} 
\usepackage{multirow} 
\usepackage{tabularx} 
\usepackage[ruled,linesnumbered]{algorithm2e} 
\usepackage{xspace} 
\usepackage{pgfplots}\pgfplotsset{compat=1.9} 
\usepackage{balance} 
\usepackage{soul} 
\usepackage{subcaption} 
\usepackage{bm} 

\usetikzlibrary{positioning}

\usepackage{xcolor}
\newcommand{\new}[1]{#1}

\newcolumntype{L}[1]{>{\raggedright\let\newline\\\arraybackslash\hspace{0pt}}m{#1}}
\newcolumntype{C}[1]{>{\centering\let\newline\\\arraybackslash\hspace{0pt}}m{#1}}
\newcolumntype{R}[1]{>{\raggedleft\let\newline\\\arraybackslash\hspace{0pt}}m{#1}}

\theoremstyle{definition}

\newcommand*\code{\texttt}
\providecommand\vec{}
\renewcommand{\vec}[1]{\ensuremath{\bm{#1}}}

\DeclareCaptionFont{xbf}{\bfseries\boldmath}
\copyrightyear{2022}
\acmYear{2022}
\setcopyright{acmlicensed}\acmConference[WSDM '22]{Proceedings of the Fifteenth ACM International Conference on Web Search and Data Mining}{February 21--25, 2022}{Tempe, AZ, USA}
\acmBooktitle{Proceedings of the Fifteenth ACM International Conference on Web Search and Data Mining (WSDM '22), February 21--25, 2022, Tempe, AZ, USA}
\acmPrice{15.00}
\acmDOI{10.1145/3488560.3498529}
\acmISBN{978-1-4503-9132-0/22/02}

\begin{document}
\fancyhead{} 

\title{Time Masking for Temporal Language Models}

\author{Guy D. Rosin}
\affiliation{
  \institution{Technion}
  \city{Haifa}
  \country{Israel}
}
\email{guyrosin@cs.technion.ac.il}
\author{Ido Guy}
\affiliation{
  \institution{Ben-Gurion University of the Negev}
  \city{Beer-Sheva}
  \country{Israel}
}
\email{idoguy@acm.org}
\author{Kira Radinsky}
\affiliation{
  \institution{Technion}
  \city{Haifa}
  \country{Israel}
}
\email{kirar@cs.technion.ac.il}


\begin{abstract}

Our world is constantly evolving, and so is the content on the web. Consequently, our languages, often said to mirror the world, are  dynamic in nature. 
However, most current contextual language models are static and cannot adapt to changes over time. 
In this work, we propose a temporal contextual language model called TempoBERT, which uses time as an additional context of texts.
Our technique is based on modifying texts with temporal information and performing time masking---specific masking for the supplementary time information. 
We leverage our approach for the tasks of semantic change detection and sentence time prediction, experimenting on diverse datasets in terms of time, size, genre, and language.
Our extensive evaluation shows that both tasks benefit from exploiting time masking.

\end{abstract}

\ccsdesc[500]{Computing methodologies~Lexical semantics}

\keywords{temporal semantics; semantic change detection; language models}

\maketitle

\section{Introduction}
\label{sec:intro}



The ability to learn from context is crucial for language understanding and modeling, as proved by the rise of contextual language models in recent years~\cite{devlin2019bert,liu2019roberta}. 
But what is context? Context has always been textual, i.e., neighboring words~\cite{mikolov2013distributed,devlin2019bert}. 
We argue there are other types of context that are valuable for language modeling, and propose to use time as context.

Our language is continuously evolving, especially for content on the web.
The rate of change of the web, language, and culture have compressed the time in which critical changes happen~\cite{adar2009web}.
Many important tasks in Natural Language Processing and Information Retrieval, such as web search, must scale not only to the number of documents, but also temporally~\cite{kanhabua2016temporal,huang2019neural,rottger2021temporal,savov2021predicting}. 
Language models (LMs) are usually pretrained on corpora derived from a snapshot of the web crawled at a specific moment in time~\cite{devlin2019bert,liu2019roberta}. 
This ``static'' nature of training prevents language models from adapting to time and generalizing temporally~\cite{rottger2021temporal,lazaridou2021pitfalls,hombaiah2021dynamic,dhingra2021time}.

\begin{figure*}
    \centering
    \small 
    \subfloat[\centering TempoBERT is trained on temporal corpora, where each sequence is prepended with temporal context information.
    \label{fig:tempobert_training}]{
    \begin{tikzpicture}
        \node (corpus) at (0, -0.13) {\includegraphics[width=27pt]{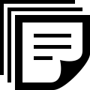}};
        \node[anchor=west] at (0.26,0.5) (sent1) {\textsc{Year: 1800}};
        \node[anchor=west] at (0.26,-0.7) (sent2) {\textsc{Year: 2020}};
        
        \node[anchor=west] (sent1_new) [right=4mm of sent1] {"\textcolor{violet}{\code{<1800>}} The mountains have an awful majesty."};
        \node[anchor=west] at (sent2 -| sent1_new.west) (sent2_new) {"\textcolor{violet}{\code{<2020>}} You look awful today."};
        
        \draw[->] (sent1) -- (sent1_new);
        \draw[->] (sent2) -- (sent2_new);
     \end{tikzpicture}
    }
    ~
    \subfloat[\centering TempoBERT can be used for inference in two modes: (1) time prediction; (2) time-dependent mask filling.
    \label{fig:tempobert_inference}]{
    \begin{tikzpicture}
        \node[anchor=west] (title1) {\textbf{Time prediction:}};
        \node[anchor=west,align=left] at (0,-1) (title2) {\textbf{Time-dependent MLM:}};
        
        \node[anchor=west] (sent2) [right=1mm of title2,yshift=2.3mm] {"\textcolor{violet}{\code{<1800>}} He has an awful [MASK]."};
        \node[anchor=west] (sent3) [right=1mm of title2,yshift=-2.3mm] {"\textcolor{violet}{\code{<2020>}} He has an awful [MASK]."};
        \node[anchor=west] at (title1 -| sent2.west) (sent1) {"[MASK] Today's weather is awful."};
        
        \node[anchor=west] (sent1_new) [right=4mm of sent1] {\textcolor{violet}{\code{<2020>}}};
        \node[anchor=west] at (sent2 -| sent1_new.west) (sent2_new) {presence};
        \node[anchor=west] at (sent3 -| sent2_new.west) (sent3_new) {temper};
        
        \draw[->] (sent1) -- (sent1_new);
        \draw[->] (sent2) -- (sent2_new);
        \draw[->] (sent3) -- (sent3_new);
    \end{tikzpicture}
    }
    \caption{Example of TempoBERT's time masking for training and inference. The word `awful' changed its meaning in the last two centuries from marvelous to disgusting.
    }
    \label{fig:tempobert}
\end{figure*}

At the heart of the masked language modeling (MLM) approach is the task of predicting a masked subset of the text given the remaining, unmasked text. 
In this work, we adapt this approach and propose a simple way to encode time into a language model. 
We explicitly model temporality by adding time information directly into texts.
This modification only yields changes in the input representations, and is thus model agnostic.
Most existing temporal models train separate models specialized to different time periods~\citep{hamilton2016diachronic}. In contrast, we create a single model that is inherently temporal.
In our approach, sequences in the training data are concatenated with a special token of the time period.
The concatenation step enables the learning of word representations that are specific to each time period.
For example, ``\code{<2021>} Joe Biden is the President of the United States'', where \code{<2021>} is a special time token added to the model's vocabulary by our framework to represent the year of 2021.
By applying this modification contextual embedding methods can utilize the temporal information attached to the input text. 
We present several masking techniques to exploit this modification when pretraining LMs, as part of (or in addition to) the standard language modeling objective of BERT~\cite{devlin2019bert}.
We refer to this process as \textit{time masking}, and to this modification applied to BERT models as \textit{TempoBERT} (Section~\ref{sec:time_masking}). 
Figure~\ref{fig:tempobert} depicts the training process of TempoBERT, along with different ways to use it for inference.

It was previously argued that the embeddings generated by BERT~\cite{devlin2019bert} are by definition dependent on the time-specific context and therefore do not require diachronic fine-tuning~\cite{martinc2020leveraging}.
In this work, we challenge this hypothesis by training the language model in a temporal fashion, using time masking. 
We experiment two tasks---semantic change detection and sentence time prediction.

Semantic change detection is the task of identifying which words undergo semantic changes and to what extent.
Most existing contextual methods detect changes by first embedding the target words in each time period, and then either aggregating them to create a time-specific embedding~\cite{martinc2020leveraging}, or computing a cluster of the embeddings for each time~\cite{giulianelli2020analysing,martinc2020capturing,montariol2021scalable,laicher2021explaining}. The embeddings or clusters are compared to estimate the degree of change between different time periods.
These approaches are similar to the traditional static word embedding models, where each word from the predefined vocabulary is presented as a unique vector, and to find the degree of change, the vectors of a word at different times are compared~\cite{hamilton2016diachronic,schlechtweg2019wind}. 
In this work, we propose a new method to detect changes, exploiting time masking (Section~\ref{sec:scd}). Through time masking, it is no longer needed to focus on the target word embedding, but rather learn the context of the time token.
TempoBERT detects changes by analyzing the distribution of the times predicted by the transformers. 
We experiment with several diverse datasets in terms of time, size, genre, and language. Our empirical results show that TempoBERT outperforms state-of-the-art methods~\cite{del2019short,schlechtweg2019wind,martinc2020leveraging,montariol2021scalable}.

The second task is sentence time prediction. 
A textual unit has two main temporal dimensions: its timestamp, or creation time, and its focus time~\citep{campos2014survey}. Both dimensions are important to meaningfully answer temporal queries in Information Retrieval systems.
Knowledge of the creation time of texts is essential for many important tasks, such as temporal search, summarization, event extraction, or temporally-focused information extraction. 
While existing approaches for these tasks assume accurate knowledge of the creation time, it is not always available, especially for arbitrary documents from the Web.
For most of the documents on the Web, the time-stamp metadata is either missing or cannot be trusted~\citep{kanhabua2008improving}. 
Thus, predicting the time of textual units (e.g., documents, posts, or sentences) based on their content is an important task~\cite{de2005t,kumar2011supervised,vashishth2018dating,savov2021predicting}.
We tackle this task as a multiclass classification problem, with classes defined as time intervals (e.g., years or decades).
In this work, we focus on dating sentences, which is a harder task than dating documents, and is a building block of the document dating task~\cite{savov2021predicting}. We use time masking to directly predict the writing time of sentences (Section~\ref{sec:stp}).
We experiment with two datasets of different granularities (i.e., years and decades) and find that TempoBERT outperforms classifiers based on GloVe~\cite{pennington2014glove} or BERT~\cite{devlin2019bert} embeddings.

Overall, our work offers the following contributions: (1) We propose a simple modification to pretraining that facilitates the acquisition of temporal knowledge through time masking; (2) We conduct evaluations on the tasks of semantic change detection and sentence time prediction that demonstrate the impact of temporally-scoped pretraining; (3) On semantic change detection, we reach state-of-the-art performance on various datasets; (4) We perform a quantitative analysis of time masking for the two evaluation tasks. We also publish our code and data.\footnote{\url{https://github.com/guyrosin/tempobert}}

\section{Related Work}

\paragraph{Temporal Language Models}
Language models are usually pretrained on corpora derived from a snapshot of the web crawled at a specific moment in time~\cite{devlin2019bert,liu2019roberta}.
This ``static'' nature of training prevents LMs from adapting to time and generalizing temporally~\cite{rottger2021temporal,lazaridou2021pitfalls,hombaiah2021dynamic,dhingra2021time}.
\citet{hombaiah2021dynamic} performed incremental training to better handle continuously evolving web content.
\citet{dhingra2021time} experimented with temporal language models for question answering. The authors focused on probing LMs for factual knowledge that changes over time, and showed that conditioning the temporal LM on the temporal context of the text data improves memorization of facts. 
Others focused on document classification by using word-level temporal embeddings~\cite{huang2019neural} and adapting pretrained BERT models to domain and time~\cite{rottger2021temporal}.

In this work, we focus on the novel concept of temporal contextual representation. 
We leverage our proposed model for the tasks of semantic change detection and sentence time prediction, which were not studied in the above papers. 
Furthermore, we learn \emph{temporal contexts} for word embeddings, introduce the concept of time masking, and demonstrate its benefits for temporal LMs.

\paragraph{Semantic Change Detection}
Semantic change detection is the task of determining whether and to what extent the meaning of a set of target words has changed over time, with the help of time-annotated corpora~\cite{kutuzov2018diachronic,tahmasebi2018survey}. This task is often addressed using distributional semantic models: time sensitive word representations 
are learned and then compared between different time periods~\cite{hamilton2016diachronic,bamler2017dynamic,dubossarsky2019time}.
\citet{gonen2020simple} used a simple nearest-neighbors-based approach to detect semantically-changed words.
\citet{del2019short} measured changes of word meaning in online Reddit communities by employing an incremental fine-tuning approach. 
\citet{montariol2021scalable} used statistical tools to detect semantic change in a scalable way using clusters.
Others attempted to simultaneously learn time-aware embeddings over all time periods and resolve the alignment problem using regularization~\cite{yao2018dynamic}, modeling word usage as a function of time~\citet{rosenfeld2018deep},  Bayesian skip-gram model with latent time series~\cite{bamler2017dynamic}, and exponential family embeddings~\cite{rudolph2018dynamic}.

In all of these methods, each word has a single representation for each time period, which limits their sensitivity and interpretability. 
Recent contextualized architectures allow for overcoming this limitation by taking sentential context into account when inferring word token representations. Indeed, such architectures were applied to diachronic semantic change detection~\cite{hu2019diachronic,giulianelli2020analysing,martinc2020leveraging,martinc2020capturing,laicher2021explaining}. While all these studies used language models by aggregating the information from the set of the token embeddings, we directly exploit their contextual properties by integrating time \emph{directly} into their training process.

\paragraph{Sentence Time Prediction}
In the last decade, the exploitation of temporal information to improve document search and exploration received a notable attention from the Information Retrieval community. Specifically, in this context, Temporal Information Extraction is crucial to support retrieval systems~\cite{campos2014survey,kanhabua2016temporal}.
Predicting the time of textual units (e.g., documents, posts, or sentences) based on their content is an important task~\cite{de2005t,kumar2011supervised,vashishth2018dating,savov2021predicting} with numerous approaches. 
Most existing work uses documents as textual units: 
One of the first studies to model temporal information for the automatic dating of documents is~\citet{de2005t}, who used a statistical LM based on word usage statistics over time. 
Several papers extended this model with temporal entropy~\cite{kanhabua2008improving}, KL divergence~\cite{kumar2011supervised}, and n-grams~\cite{jatowt2017interactive}. 
\citet{niculae2014temporal} tackled this task as a ranking problem. 
As for deep neural models for document dating, \citet{vashishth2018dating} used a Graph Convolution Network together with the syntactic and temporal structure of the document, and \citet{ray2018ad3} used an attention-based neural model.
Recently, \citet{savov2021predicting} used Ordinal Classification to predict the time of each sentence individually and then aggregated the scores to document level.

In this work, we focus on dating sentences. Sentences are often significantly shorter than documents, making this task potentially harder than dating documents, due to the limited context. In addition, sentence dating can be used as a building block of document dating, as was done in~\cite{savov2021predicting}. We use time masking to directly predict the writing time of sentences.

\section{Time Masking}
\label{sec:time_masking}

In this work, we use bidirectional language models (also called ``contextual language models''). 
These models represent words as a function of the entire context of a unit of text (e.g., sentence or paragraph), and not only conditioned on previous words as in traditional unidirectional models.
Formally, given an input sequence of tokens $\vec{w} = [w_1, w_2, \dots , w_n]$ and a position $1 <= i <= n$, bidirectional LMs estimate $p(w_i) = p(w_i \mid w_1, \dots , w_{i-1}, w_{i+1}, \dots , w_n)$ using the left and right context of that word.
To estimate this probability, BERT~\cite{devlin2019bert} samples positions in the input sequence randomly and learns to fill the word at the masked position. To this end, a Transformer architecture is employed.

To adapt language models to time, we propose to directly encode temporality information in the training process.
We design a modification to the training process designed specifically for contextual representations: rather than including time information for each word as in existing approaches (e.g., temporal referencing~\cite{dubossarsky2019time}), we concatenate a special time token to each textual sequence. 
For example, ``<2021> Joe Biden is the President of the USA'', where ``<2021>'' is a special time token added to the model's vocabulary by our framework to represent the year of 2021.
By performing this modification at the sequence level instead of at the token level, \new{we avoid adding time-specific variants of each token to the vocabulary, thus increasing its size significantly. 
In addition, contextual embedding methods can utilize this temporal information attached to the input text.}
The time token denotes the time at which the sequence was written or a point in time at which its assertion is valid.

In addition to encoding time into the model and thus making the model temporal, time masking enables predicting the writing time of texts.
The same way prompts~\cite{petroni2019language,jiang2020can} exploit the masked language model to fill mask tokens for the task of question answering, we can fill mask time tokens to predict the time of the sequence.
We exploit this feature in both tasks we experiment with in this work: semantic change detection and sentence time prediction.
For example, to predict the time of the sentence ``Joe Biden is the President of the USA'', we prefix it with a mask time token: ``\code{[MASK]} Joe Biden is the President of the USA'' and feed it into the model. The model is likely to fill the mask token with a time token, since it was pretrained on such sentences. Thus, the expected token is the relevant time token to Biden's presidency, e.g., ``\code{<2021>}''. 

Formally, given a sequence \vec{w} and a time $t$, we denote $w_0 = t$ for clarity and prepend the time token $t$ to \vec{w}.
Now the inputs to the model are a sequence $\vec{w} = [t, w_1, w_2, \dots , w_n]$ and a position $0 <= i <= n$.
Let us analyze the two options for $i$: 
(1) If $1 <= i <= n$, then $w_i$ is one of the original tokens in the sentence. In this case, $p(w_i) = p(w_i \mid t, w_1, \dots , w_{i-1}, w_{i+1}, \dots , w_n)$, i.e.,  $p(w_i)$ is computed based on both the textual context and the temporal context.
(2) If $i = 0$, then $w_i$ is $t$. In this case, $p(w_i) = p(t) = p(t \mid w_1, \dots , w_n)$, i.e., we predict the time $t$ by the whole input text.

As for the granularity of $t$, different values can be used, according to the use case.
In this work, for the sentence time prediction task (Section~\ref{sec:stp}), we experiment with granularity of a year and of a decade. 
For the semantic change detection task (Section~\ref{sec:scd}), the corpora are divided into two time buckets, so we use two time tokens, one for each bucket (i.e., \code{<1>} and \code{<2>}).

Throughout this paper, we use pretrained BERT-base~\cite{devlin2019bert} (with 12 layers, 768 hidden dimensions, and 110M parameters)\footnote{We use Hugging Face's implementation of BERT: \url{https://github.com/huggingface/transformers}.}, and then post-pretrain it on the temporal corpora with time masking, as described above. We refer to the resulting model as \textit{TempoBERT}.

Finally, time masking can be performed in two ways: 
The straightforward one is to use the standard MLM, where time tokens are treated like any other token for the masking process.
But we can also use a \textit{separate} masking process for time tokens. In this case, we choose time tokens with a custom probability $p_{TM}$; a chosen time token is then either masked (i.e., replaced with the mask token \code{[MASK]}) 80\% of the time, replaced by a random time token 10\% of the time, or kept unchanged 10\% of the time (we follow~\citet{devlin2019bert} for the 80-10-10 probabilities). 
See Section~\ref{sec:time_masking_analysis} for an analysis of the different settings.

\section{Applications for Temporal Contextual Representations}

\subsection{Semantic Change Detection}
\label{sec:scd}
\new{Words can change their meaning, gain or lose senses over time. 
Knowing whether, and to what degree, a word has changed is key to various tasks}, e.g., aiding in understanding historical documents, searching for relevant content, or historical sentiment analysis.
In semantic change detection~\cite{kutuzov2018diachronic,tahmasebi2018survey}, the task is to rank a set of target words according to their degree of semantic change between two time periods $t_1$ and $t_2$.

We propose two scoring functions that utilize TempoBERT for this task (see Section~\ref{sec:analysis_long_short_term} for an analysis of them): given a target word $w$, they approximate the semantic change it underwent between $t_1$ and $t_2$.

\paragraph{temporal-cosine}
We generate time specific representations of words and compare them to detect semantic changes. This method was used by \citet{martinc2020leveraging}. It is model agnostic and does not explicitly utilize time masking.
In detail, we sample $n$ sentences containing the target word $w$ from each time period $t$ and feed them into the model.
For each of the sentences, we create a sequence embedding by extracting their last $h$ hidden layers and averaging them. We then extract the embedding of the specific target word. This is the contextual token embedding. 
Note that these representations vary according to the context in which the token appears, meaning that the same word has a different representation in each specific context (sequence).
Following, the resulting embeddings are aggregated at the token level and averaged, in order to get a time specific representation for $w$ for each time $t$, denoted by $e_t(w)$.
Finally, we estimate the semantic change of $w$ by measuring the cosine distance (\textit{cos\_dist}) between two time specific representations of the same token:
\begin{align*}
\textit{score}(w) = \textit{cos\_dist}(e_{t_1}(w), e_{t_2}(w))
\end{align*}

\paragraph{time-diff}
This method is tailored to TempoBERT. We use our model's ability to predict the time of sentences and identify semantic change by looking at the distribution of the predicted times.
Specifically, we sample $n$ sentences containing the target word $w$ from each time period and add them all to a set of sentences for $w$, denoted by $S_w$.
For each sentence $s \in S_w$, we prepend a mask token and feed it into the model for the task of MLM.
The model fills the mask token with the most appropriate token, which in our case is a time token (since during our post-pretraining, the model was trained on sentences starting with a time token).
This produces a predicted time for the sentence.
Let $p_{t}(s)$ be the probability of time $t$ to be predicted for $s$.
Consider the distribution of these predicted probabilities (for each of the time periods).
Intuitively, a uniform distribution means those sentences could have been written at any time period, which means there was no semantic change. A non-uniform distribution implies there was a change that caused the sentences to be written in a specific time period.
In our case, since there are just two time periods, we simply use the absolute difference of the probabilities $|p_{t_1}(s) - p_{t_2}(s)|$, so the semantic change score of the target word is set to the average difference of the predicted time probabilities:
\begin{align*}
\textit{score}(w) = \frac{1}{|S_w|} \sum_{s \in S_w} |p_{t_1}(s) - p_{t_2}(s)|
\end{align*}

\subsection{Sentence Time Prediction}
\label{sec:stp}
As another evaluation of TempoBERT's generalization potential, we examine its use for the task of sentence time prediction; it is a multiclass classification problem, where the goal is to predict the time of a sentence~\cite{de2005t,kumar2011supervised,vashishth2018dating,savov2021predicting}.
This task is a building block of the document dating task~\cite{savov2021predicting}.
Tackling this task with TempoBERT is straightforward: given an input sentence, we prepend a mask token and predict it.
In more detail, we feed the prepended sentence into the model for the task of MLM.
The model fills the mask token with the most appropriate token, which in our case is a time token (since during our post-pretraining, the model was trained on sentences starting with a time token).
This produces a predicted time for the sentence.


\section{Datasets}
\label{sec:datasets}

\subsection{Semantic Change Detection Datasets}
For semantic change detection, we experiment with several diachronic data sources, covering a variety of genres, time periods, languages, and sizes.
We use the LiverpoolFC corpus~\cite{del2019short} and the SemEval-2020 Task 1 corpora~\cite{schlechtweg2020semeval} (English and Latin). 
Each dataset is split into two time periods; see Table~\ref{tab:scd_datasets} for the datasets' statistics.

\paragraph{LiverpoolFC~\cite{del2019short}}
The LiverpoolFC corpus contains 8 years of Reddit posts from the Liverpool Football Club subreddit. It was created for the task of short-term meaning shift analysis in online communities. 
The corpus is in English and split into two time periods from 2011-2013 and 2017. 
We follow~\citet{martinc2020leveraging} and only conduct light text preprocessing on the LiverpoolFC corpus, where we remove URLs.

Semantic change evaluation dataset: \citet{del2019short} published a set of 97 words from the LiverpoolFC corpus, manually annotated with semantic shift labels by the members of the LiverpoolFC subreddit. 
\new{Each word was assigned a graded label (between 0 and 1) according to its degree of semantic change.
The average of these judgements is used as a gold standard semantic shift index.}

\paragraph{SemEval-2020 Task 1 datasets~\cite{schlechtweg2020semeval}}
The SemEval-2020 Task 1 deals with the semantic change detection task. It introduced datasets in four different languages. In this work, we use the English and Latin datasets. They are all long-term, in contrast to the LiverpoolFC dataset: the English dataset spans two centuries, and the Latin dataset spans more than 2000 years. 

Semantic change evaluation dataset: The task contains two subtasks: binary classification of whether a word sense has been gained or lost (subtask 1) and ranking a list of words according to their degree of semantic change (subtask 2). 
\new{We use the data for subtask 2, which is a set of target words and their graded labels.}
The target words are balanced for part of speech (POS) and frequency.
\new{For the English dataset, we remove POS tags from both the corpus and the evaluation set (as done in previous work~\cite{montariol2021scalable}).}

\begin{table*}
\caption{Semantic change detection datasets composition. Each dataset is split into two time periods, denoted by C1 and C2.}
\label{tab:scd_datasets}
\centering
\begin{tabular}{@{}llllllll@{}}
    \toprule
    Dataset & Labelled Words & C1 Source & C1 Time Period & C1 Tokens & C2 Source & C2 Time Period & C2 Tokens\\
    \midrule
    LiverpoolFC & 97 & Reddit & 2011--2013 & 8.5M & Reddit & 2017 & 11.9M\\
    SemEval-English & 37 & CCOHA & 1810--1860 & 6.5M & CCOHA & 1960--2010 & 6.7M\\
    SemEval-Latin & 40 & LatinISE & -200--0 & 1.7M & LatinISE & 0--2000 & 9.4M\\
    \bottomrule
\end{tabular}
\end{table*}

\subsection{Sentence Time Prediction Datasets}

We use the New York Times archive 
with 40 years of news articles (from 1981 to 2020, 11 GB of text in total).
We use two variants of this corpus (Table~\ref{tab:stp_datasets}); one with a granularity of decades (\textit{NYT-decades}) and the other with a granularity of individual years (\textit{NYT-years}).
For both of them, we downsample the corpus to 10k sentences per year (1.4M tokens in average, around 1.4 MB of text) to ease computation.
The year-granularity dataset contains 40 years, so we have 40 classes for our classification task.
To create the decade-granularity dataset, we merge each decade. The resulting dataset contains 4 decades (i.e., four classes), each is composed of 14 MB of text. 
Each row in the two datasets is composed of a sentence and its associated time (year or decade).
For the evaluation, we randomly split both datasets to train and test sets using an 80-20 ratio.
\begin{table*}
\caption{Sentence time prediction datasets composition.}
\label{tab:stp_datasets}
\centering
\begin{tabular}{@{}llllll@{}}
    \toprule
    Dataset & Source & Time Period & Classes & Avg. Tokens per Period & Avg. Tokens per Sentence\\
    \midrule
    NYT-decades & NYT Archive & 1981--2020 & 4 & 14M & 22.2\\
    NYT-years & NYT Archive & 1981--2020 & 40 & 1.4M & 22.2\\
    \bottomrule
\end{tabular}
\end{table*}

\section{Experimental Setup}
\label{sec:evaluation}

\subsection{Semantic Change Detection Setup}
\label{sec:scd_setup}

\paragraph{Baseline methods}
We use five baseline methods for semantic change detection:
(1) \citet{del2019short} was the first paper to address the notion of short-term semantic change detection. They use a Skip-gram with Negative Sampling model (SGNS)~\cite{mikolov2013distributed} and employ an incremental model fine-tuning approach; the model is first trained on a large Reddit corpus of around 900M tokens (a random crawl from Reddit in 2013) for the initialization of the word vectors, and then fine-tuned on the LiverpoolFC corpus.
(2) \citet{schlechtweg2019wind} is the state-of-the-art semantic change detection method employing non-contextual word embeddings: the Skip-gram with Negative Sampling model is trained on two periods independently and aligned using Orthogonal Procrustes. Cosine distance is used to compute the semantic change.
(3) \citet{gonen2020simple}, who use SGNS embeddings along with a simple detection method. For each period, a word is represented by its top nearest neighbors according to cosine distance. Semantic change is then measured as the size of the intersection between the nearest neighbors lists of two periods.
(4) \citet{martinc2020leveraging} were one of the first to use BERT for semantic change detection, by creating time-specific embeddings of words, and comparing between different times by calculating the cosine distance between the averaged embeddings. Notice, unlike our work they consider time-specific context rather than capture temporal context. They focused on short-term semantic change as well. Their model performs slightly worse than \citet{del2019short}, but it is more general and scalable.
(5) \citet{montariol2021scalable} generate a set of contextual embeddings using BERT for each word. These representations are clustered, and the derived cluster distributions are compared across time slices. Finally, words are ranked according to the distance measure, assuming that the ranking resembles a relative degree of usage shift. 
We use their best performing method as reported in~\cite{montariol2021scalable}, which uses affinity propagation for clustering word embeddings and Wasserstein distance as a distance measure between clusters. 
For the LiverpoolFC dataset, we experiment with all the variants proposed in~\cite{montariol2021scalable} and choose the best performing one, which uses k-means with $k=7$ for clustering embeddings, and Wasserstein distance as a distance measure between clusters.

Finally, existing methods for this task that concatenate time information to individual words (i.e., Temporal Referencing~\cite{dubossarsky2019time} and Word Injection~\citep{schlechtweg2019wind}) were found~\cite{schlechtweg2019wind,schlechtweg2020semeval} to be outperformed by~\citet{schlechtweg2019wind}, so we leave them out of this evaluation.

\paragraph{Our method}
For each dataset, we train a TempoBERT model using a pretrained case-insensitive BERT-base model.\footnote{For English: bert-base-uncased from the Hugging Face library, and for Latin: latin-bert from \url{https://github.com/dbamman/latin-bert}.}
Since the vocabulary of this model does not contain all the target words in the evaluation datasets, we add all target words to TempoBERT's vocabulary (to avoid the tokenizer splitting any occurrences of the target words to subwords).

\paragraph{Metrics}
Semantic change detection performance is measured by the correlation between the semantic shift index (i.e., the ground truth) and the model's semantic shift assessment for each word in the evaluation set. 
Previous work used different correlation measures for different datasets; for the LiverpoolFC dataset, Pearson's $r$ was used~\cite{del2019short,martinc2020leveraging}, while for the SemEval-2020 datasets it was Spearman's rank-order correlation coefficient $\rho$~\cite{schlechtweg2020semeval,montariol2021scalable}. 
The difference between them is that Spearman's $\rho$ only considers the order of the words, while Pearson's $r$ considers the actual predicted values.
In our evaluation, we make an effort to evaluate our methods and the baselines using both Pearson and Spearman correlation, to make the evaluation as comprehensive as possible. 
There were some cases where we could not reproduce the original authors' results; in such cases we opted to report only the original result.

\subsection{Sentence Time Prediction Setup}
\label{sec:stp_setup}

\paragraph{Baseline Methods}
We use the following baselines:
(1) CNN~\cite{kim2014convolutional}: a convolutional neural network loaded with pretrained GloVe~\cite{pennington2014glove} vectors.
(2) Bi-LSTM + Attention~\cite{zhou2016attention}: an attention-based bidirectional Long Short-Term Memory Network loaded with pretrained GloVe vectors.
(3) Linear GloVe: a linear model with one hidden layer of 10 units and a softmax, loaded with pretrained GloVe vectors.
(4) Boosting methods based on BERT embeddings (extracted using BERT's last four hidden layers): XGBoost and CatBoost.\footnote{\url{https://github.com/dmlc/xgboost}, \url{https://github.com/catboost/catboost}}

\paragraph{Our method}
For each dataset, we train a TempoBERT model using a pretrained uncased BERT-base model with time masking probability $p_{TM}=0.9$. 

\paragraph{Metrics}
As sentence time prediction is a multiclass classification task, we report accuracy and macro-F1 scores on the test sets. 

\subsection{Implementation Details}
\label{sec:impl}
Due to limited computational resources, we train our models with a maximum input sequence length of 128 tokens. 
To train TempoBERT models, we tune the following hyperparameters for each dataset: learning rate $l \in \{1\text{e-}7, 1\text{e-}6, 1\text{e-}5, 1\text{e-}4, 2\text{e-}4, 3\text{e-}4\}$ and number of epochs $e \in \{1, 2, 5, 10, 20\}$.
To use TempoBERT for semantic change detection, we use a maximum number of sentences $n=200$, to ease computation (previous work used all available sentences~\cite{martinc2020leveraging}).
Specifically for the temporal-cosine method, we tune the number of last hidden layers to use for embedding extraction $h \in \{1, 4, 12\}$.
All experiments were run using four NVIDIA GeForce RTX 2080 Ti GPUs. 
Runtime varied by experiment; training our best TempoBERT models on each of the datasets took between 30 minutes and 4 hours.

For fine-tuning BERT on labelled data, we train for three epochs with a batch size of 32, which corresponds to the default settings recommended by \citet{devlin2019bert}.

\section{Results}
\label{sec:results}

\subsection{Semantic Change Detection Results}
\label{sec:results_scd}

Table~\ref{tab:scd} shows the results for semantic change detection on the LiverpoolFC and SemEval datasets.
As can be seen, TempoBERT outperforms all the baselines using both Pearson and Spearman correlation with significant correlations ($p<0.002$), except for the SemEval-English dataset, where it is outperformed by~\cite{montariol2021scalable} by 0.028, although TempoBERT got a higher Spearman correlation by 0.03.
For the LiverpoolFC dataset, we observe a strong correlation (Pearson's $r=0.637$ and Spearman's $\rho=0.620$) along with a very large performance gap compared to the baselines (which achieve around $0.490$ for both measures, at max). 
For the SemEval corpora, we observe a moderate correlation (around 0.47--0.54) and a relatively smaller (but still considerable) 
gap compared to the baselines.
The difference originates from the fact that the LiverpoolFC corpus is short-term, i.e., it contains texts from just a few years, in contrast to the SemEval corpora, which span centuries. We further analyze this phenomenon 
in Section~\ref{sec:analysis_long_short_term}.

\begin{figure}
\centering
\includegraphics[width=0.99\linewidth]{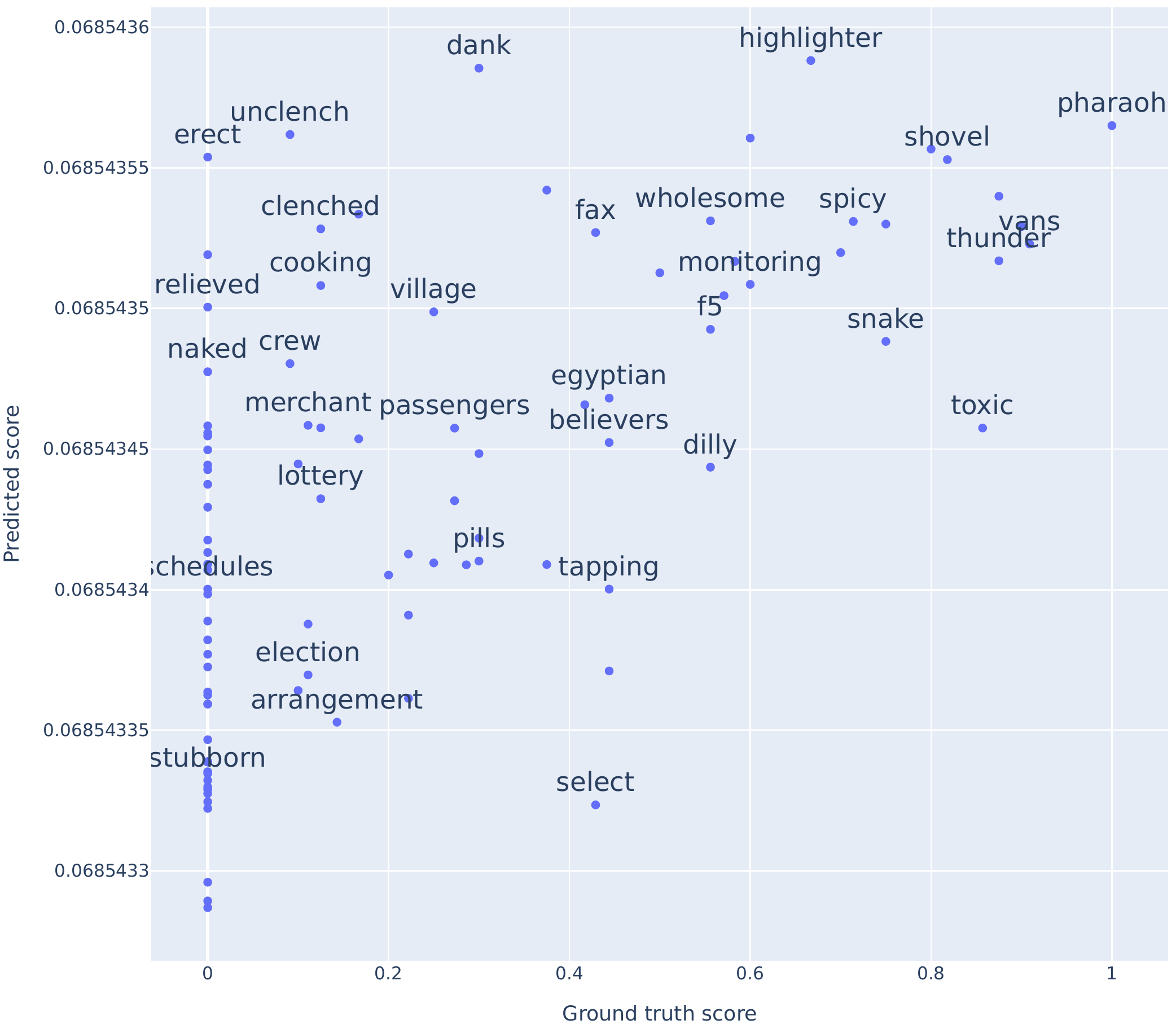}
\caption{\label{fig:liverpool_correlations}Semantic change detection on the LiverpoolFC dataset: ground truth scores vs. TempoBERT scores (Pearson's $r = 0.637$).}
\end{figure}

Figure~\ref{fig:liverpool_correlations} shows the Pearson correlation between TempoBERT's scores and the ground truth scores for the LiverpoolFC dataset. The correlation is strong ($0.637$), so the number of false positives and false negatives is considered small; there are around 10 false positive words (top-left corner), while only one false negative (bottom-right corner). Thus, while we successfully detect changed words, future work should focus on better detecting unchanged words.

\begin{table*}
\caption{Semantic change detection results on LiverpoolFC, SemEval-English, and SemEval-Latin.}
\label{tab:scd}
\centering
\begin{tabular}{@{}lllllllll@{}}
    \toprule
    \multirow{2}[3]{*}{Method} & \multicolumn{2}{c}{LiverpoolFC} & \multicolumn{2}{c}{SemEval-Eng} & \multicolumn{2}{c}{SemEval-Lat}\\
    \cmidrule(lr){2-3} \cmidrule(lr){4-5} \cmidrule(lr){6-7} &
    Pearson & Spearman & Pearson & Spearman & Pearson & Spearman\\
    \midrule
    \citet{del2019short} & 0.490 & -- & -- & -- & -- & --\\
    \citet{schlechtweg2019wind} & 0.428 & 0.425 & 0.512 & 0.321 & 0.458 & 0.372\\
    \citet{gonen2020simple} & -- & -- & 0.504 & 0.277 & 0.417 & 0.273\\
    \citet{martinc2020leveraging} & 0.473 & 0.492 & -- & 0.315 & -- & 0.496\\
    \citet{montariol2021scalable} & 0.378 & 0.376 & \textbf{0.566} & 0.437 & -- & 0.448\\
    TempoBERT & \textbf{0.637} & \textbf{0.620} & 0.538 & \textbf{0.467} & \textbf{0.485} & \textbf{0.512}\\
    \bottomrule
\end{tabular}
\end{table*}

\subsubsection{Short-term versus long-term semantic change}
\label{sec:analysis_long_short_term}
As shown in Table~\ref{tab:scd}, TempoBERT is particularly effective on short-term corpora. 
TempoBERT encodes time information at the sentence level. As a result, its strength lies in distinguishing between sentences of different times (see the evaluation on sentence time prediction in Section~\ref{sec:stp}). 
For long-term corpora, the writing style between different times (e.g., centuries) is very different; this makes the time information encoded in TempoBERT less valuable.

In long-term periods, there are major difference in writing style, which makes the prediction relatively easier to make.
In short time spans, the temporal information encoded into the texts by TempoBERT helps it predict the time accurately and without direct supervision. To demonstrate the difference between short and long term corpora, we include in our evaluation two recent methods that were originally evaluated only on long-term corpora~\cite{schlechtweg2019wind,montariol2021scalable}. As showed in Table~\ref{tab:scd}, these two methods achieved lower results compared to both TempoBERT and the two existing methods for short-range semantic change detection~\cite{del2019short,martinc2020leveraging}.

Finally, Table~\ref{tab:scd_methods_comparison} shows the comparison results of the two flavors of TempoBERT for semantic change detection, i.e., time-diff and temporal-cosine, for all the corpora, measured by both Pearson and Spearman correlation.
We can clearly see that time-diff outperforms temporal-cosine on the short-term LiverpoolFC corpus, while it is outperformed on the long-term SemEval corpora.
This further strengthens our hypothesis, as time-diff is based on sentence time prediction, while temporal-cosine is focused on the target word and is thus less affected by context differences (e.g., writing style).


\begin{table}
\caption{Semantic change detection results comparing the two TempoBERT methods time-diff and temporal-cosine on Pearson's $r$ and Spearman's $\rho$, on LiverpoolFC, SemEval-English, and SemEval-Latin.}
\label{tab:scd_methods_comparison}
\centering
\setlength{\tabcolsep}{0.45em}
\begin{tabular}{@{}lcccccccc@{}}
    \toprule
    \multirow{2}[3]{*}{Method} & \multicolumn{2}{c}{LiverpoolFC} & \multicolumn{2}{c}{SE-Eng} & \multicolumn{2}{c}{SE-Lat}\\
    \cmidrule(lr){2-3} \cmidrule(lr){4-5} \cmidrule(lr){6-7} & 
    $r$ & $\rho$ & $r$ & $\rho$ & $r$ & $\rho$\\
    \midrule
    time-diff & \textbf{0.637} & \textbf{0.620} & 0.208 & 0.381 & 0.385 & 0.425\\
    temporal-cosine & 0.371 & 0.451 & \textbf{0.538} & \textbf{0.467} & \textbf{0.485} & \textbf{0.512}\\
    \bottomrule
\end{tabular}
\end{table}


\subsubsection{Time masking experiments}
\label{sec:time_masking_analysis}
As described in Section~\ref{sec:time_masking}, time masking can be performed in two ways: (1) As part of the standard MLM (i.e., the time tokens are treated like any other token); (2) As a separate masking process, where time tokens are masked with a custom masking probability $p_{TM}$. 

In this section, we experiment with these time masking techniques: we train TempoBERT models on the LiverpoolFC corpus, each with a different method, and compare the results for the task of semantic change detection. 
We train all models using a 3e-4 
learning rate for just 5 epochs (39K steps), due to computational resource limits.

\begin{figure}
\centering
\includegraphics[width=0.8\linewidth]{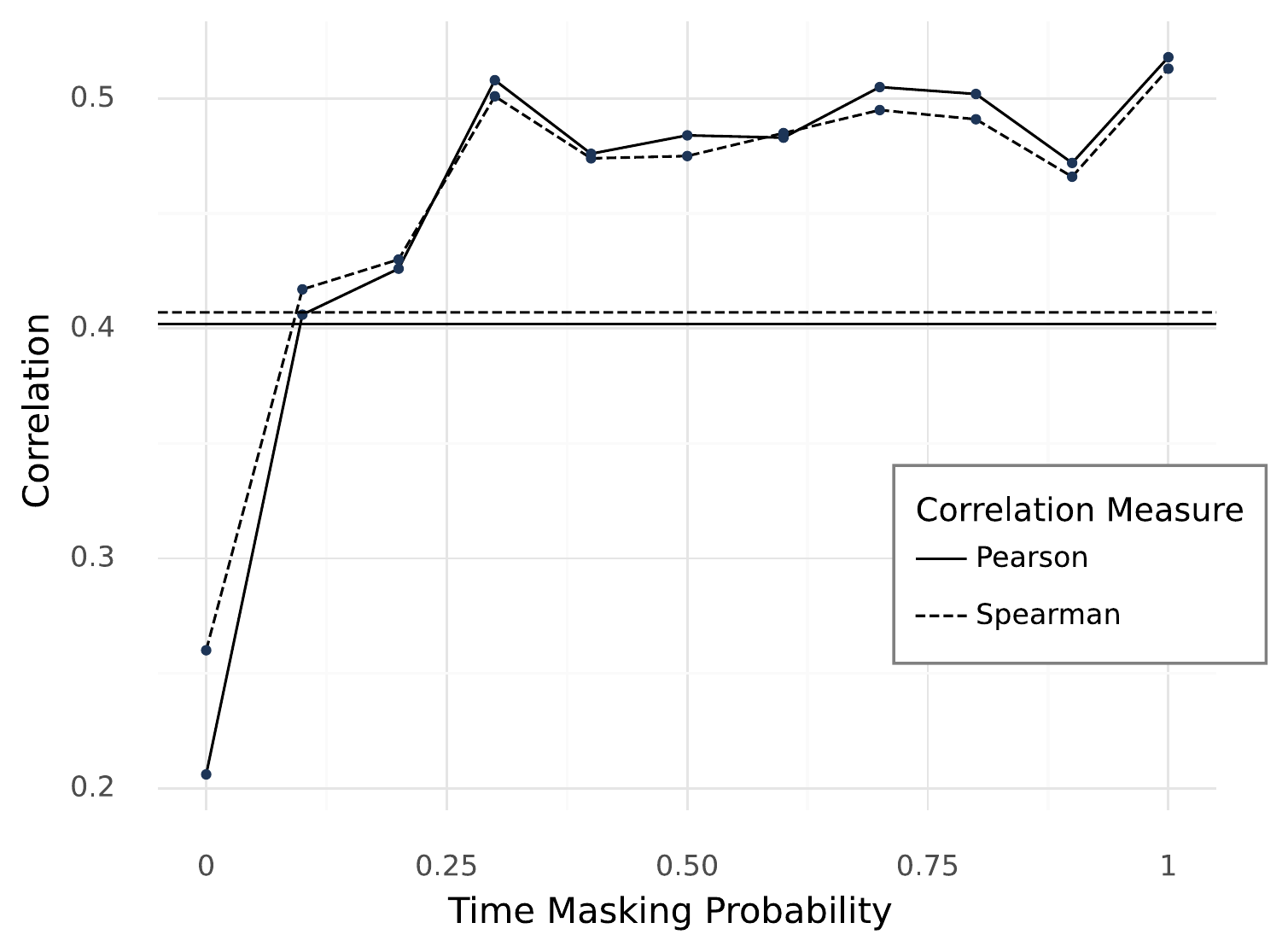}
\caption{\label{fig:time_masking_plot_scd}Time masking probability vs correlation 
for semantic change detection. The horizontal lines show the correlations for time masking as part of the standard MLM.}
\end{figure}

Figure~\ref{fig:time_masking_plot_scd} shows the comparison results, where the points show the Pearson and Spearman correlation for different time masking probabilities ($p_{TM} \in \{0.0, 0.1, \dots , 0.9, 1.0\}$), and the horizontal lines show the correlations for time masking performed as part of the standard MLM. 
We observe a positive trend from $p_{TM}=0$ to $0.3$. In particular, for $p_{TM}>=0.1$, TempoBERT benefits from custom time masking compared to standard MLM.
In addition, when training the model without time masking at all (i.e., $p_{TM}=0$), performance is much lower, as expected, with correlation around $0.2$--$0.26$.




\subsubsection{Comparison of BERT model sizes}
\label{sec:model_size_analysis}
TempoBERT is based on a pretrained BERT model with 768 hidden dimension size and 12 transformer layers. This is the most commonly used setting of BERT and is called BERT-base. 
In this section, we train TempoBERT-tiny, a variant of TempoBERT that is based on a much smaller pretrained variant of BERT, called BERT-tiny.
BERT-tiny has just $4\%$ of the parameters of BERT-base: its hidden dimension is 128, and it contains only 2 transformer layers.

Table~\ref{tab:scd_model_size_comparison} shows the comparison results, where we compare TempoBERT with TempoBERT-tiny, the best baseline for each corpus (i.e., \citet{del2019short} for LiverpoolFC, and \citet{montariol2021scalable} for SemEval-English), and a tiny variant of the best BERT-based baselines (i.e., \citet{martinc2020leveraging} for LiverpoolFC, and \citet{montariol2021scalable} for SemEval-English). 
We evaluate only on the English datasets because a pretrained BERT-tiny model is currently available only for English. 
As expected, TempoBERT-tiny achieves significant ($p<1\text{e-}5$) but lower results compared to the standard TempoBERT, but the interesting result is that even TempoBERT-tiny outperforms the best baseline on the LiverpoolFC dataset. 
On SemEval-English it is outperformed but by a small margin (0.09).
In addition, we observe relatively worse performance by the two tiny baselines: \citet{martinc2020leveraging} deteriorated from 0.473 to 0.300, and \citet{montariol2021scalable} declined from 0.437 to 0.376. Compared to them, the performance decline from TempoBERT to TempoBERT-tiny is much smaller (0.637 to 0.561, and 0.467 to 0.427).
To conclude, the tiny version of TempoBERT shows comparable or better results to the best baselines. 
We hypothesize that to understand time there is no need to use extremely large models. 
This is encouraging for both researchers possessing limited computational resources and for Earth. 
In addition, the tiny version of TempoBERT is relatively stronger compared to tiny versions of the baselines. 

\begin{table}
\caption{Results for semantic change detection for models with different sizes on LiverpoolFC (measured using Pearson) and SemEval-English (measured using Spearman).}
\label{tab:scd_model_size_comparison}
\centering
\begin{tabular}{@{}llll@{}}
    \toprule
    Method & LiverpoolFC & SemEval-Eng\\
    \midrule
    Tiny baseline & 0.300 & 0.376\\
    Best baseline & 0.490 & 0.436\\
    \addlinespace
    TempoBERT-tiny & 0.561 & 0.427\\
    TempoBERT & \textbf{0.637} & \textbf{0.467}\\
    \bottomrule
\end{tabular}
\end{table}

\subsection{Sentence Time Prediction Results}
\label{sec:results_stp}


Table~\ref{tab:stp} shows the results for sentence time prediction on the two NYT datasets. 
The two corpora show similar results: TempoBERT outperforms all the baselines. On NYT-decades, the accuracy gap between TempoBERT and the baselines is $4\%$--$8\%$, while on NYT-years the gap is around $2\%$--$3\%$. The difference between the two corpora is reasonable, since NYT-years is a much harder task (with 40 classes compared to 4 classes of NYT-decades): we observe x5--x8 higher accuracy for the evaluated methods on NYT-decades compared to NYT-years.

\begin{table}
\caption{Sentence time prediction results on NYT (4 decades) and NYT (40 years).}
\label{tab:stp}
\centering
\begin{tabular}{@{}lllll@{}}
    \toprule
    \multirow{2}[3]{*}{Method} & \multicolumn{2}{c}{NYT-decades} & \multicolumn{2}{c}{NYT-years}\\
    \cmidrule(lr){2-3} \cmidrule(lr){4-5} & 
    Acc & Mac-F1 & Acc & Mac-F1\\
    \midrule
    CNN & 46.73\% & 15.53\% & 6.44\% & 0.39\%\\
    Bi-LSTM + Attn & 48.62\% & 16.01\% & 6.69\% & 1.13\%\\
    Linear GloVe & 47.18\% & 15.71\% & 7.35\% & 0.56\%\\
    XGBoost BERT & 44.83\% & 43.75\% & 5.95\% & 5.77\%\\
    CatBoost BERT & 46.41\% & 45.14\% & 6.53\% & 6.30\%\\
    TempoBERT & \textbf{52.71\%} & \textbf{51.28\%} & \textbf{9.24\%} & \textbf{8.85\%}\\
    \bottomrule
\end{tabular}
\end{table}

\subsubsection{Time masking experiments}
\label{sec:time_masking_analysis_stp}
In this section, we experiment with various ways to perform time masking (Section~\ref{sec:time_masking}): we train TempoBERT models on the NYT-decades corpus, each with a different time masking method, and compare the results for the task of sentence time prediction. We trained all models for 10 epochs (16K steps), using a 2e-4 learning rate.

\begin{figure}
\centering
\includegraphics[width=0.7\linewidth]{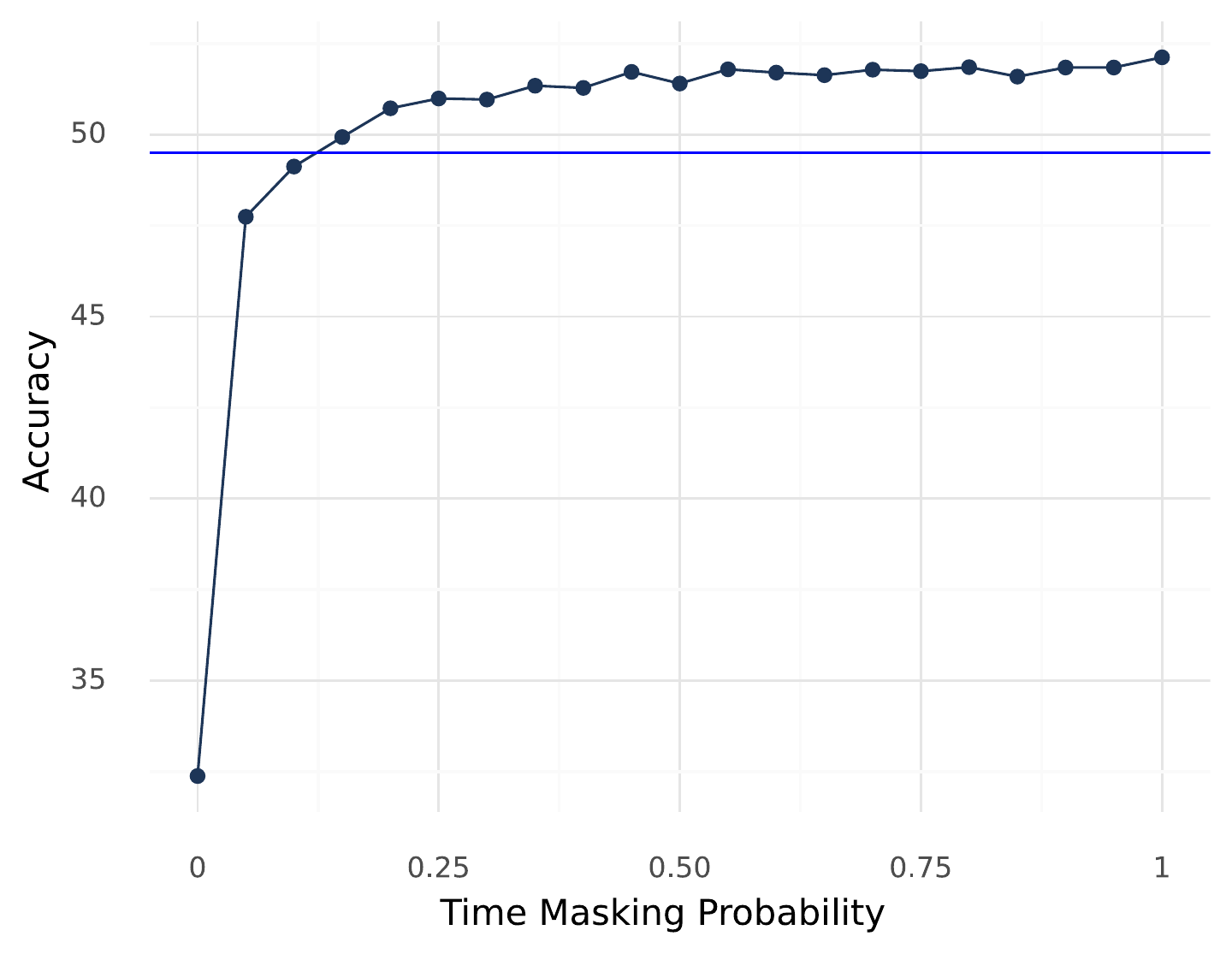}
\caption{\label{fig:time_masking_plot}Accuracy vs time masking probability for sentence time prediction. The horizontal line shows the accuracy for time masking as part of the standard MLM.}
\end{figure}

Figure~\ref{fig:time_masking_plot} shows the comparison results, where the points show the accuracy for different time masking probabilities ($p_{TM} \in \{0.0, 0.05, \\ \dots, 1.0\}$), and the horizontal line shows the accuracy for time masking performed as part of the standard MLM. 
First, we observe a positive trend, i.e., TempoBERT improves for sentence time prediction as it performs more time masking. 
This is reasonable, as the time masking itself is some kind of supervision for this task.
Second, when training the model without time masking at all (i.e., $p_{TM}=0$), performance is much lower, as expected, with 32.38\% accuracy. 
This is reasonable, as the training process in this case does not involve predicting time tokens at all, i.e., the model does not get any supervision for this task (i.e., zero-shot).
Finally, when performing time masking as part of the standard MLM, the model achieves 49.50\% accuracy. This is worse than performing time masking as a custom masking process with $p_{TM}>=0.1$.

\subsubsection{Comparison to Fine-tuned BERT}
The sentence time prediction task is a supervised task. As such, our masking approach can be compared to fine-tuning. 
It has been argued that various types of masking can be equivalent to fine-tuning in terms of performance for supervised tasks, while being more efficient~\cite{zhao2020masking}. 
We wish to examine the different parameters of fine-tuning on the performance of the sentence time prediction task to shed more light on when masking should be applied compared to fine-tuning.
Specifically, we compare TempoBERT to a BERT model fine-tuned for the task of sentence time prediction. 
Note that the supervision TempoBERT receives in its training process is much limited compared to fine-tuned BERT; fine-tuned BERT is trained directly on the sentence time prediction task, whereas TempoBERT is a general temporal language model that is trained on token time prediction as a small part of its pretraining process.

We experiment with different fine-tuned BERT classifiers, trained on different sizes of data.
Table~\ref{tab:stp_finetuned_bert_results} shows the comparison results.
TempoBERT reaches similar performance as the fine-tuned BERT at 50\% of the data. 
This shows most of the information of the time can be obtained from learning on only half the corpus but continues improving as data is available. 
When reaching 100\% of the data as compared to 50\% of it, we see a relatively smaller improvement.

\begin{table}
\caption{Comparison of different fine-tuned BERT variants for sentence time prediction on NYT (4 decades) and NYT (40 years).}
\label{tab:stp_finetuned_bert_results}
\centering
\begin{tabular}{lllll}
    \toprule
    \multirow{2}[3]{*}{Method} & \multicolumn{2}{c}{NYT-decades} & \multicolumn{2}{c}{NYT-years}\\
    \cmidrule(lr){2-3} \cmidrule(lr){4-5} & 
    Acc & Mac-F1 & Acc & Mac-F1\\
    \midrule
    Fine-tuned BERT (10\%) & 49.81\% & 48.91\% & 6.38\% & 5.33\%\\
    Fine-tuned BERT (25\%) & 51.43\% & 50.71\% & 7.14\% & 6.47\%\\
    Fine-tuned BERT (50\%) & 52.98\% & 52.37\% & 8.71\% & 8.40\%\\
    Fine-tuned BERT (100\%) & \textbf{53.08}\% & \textbf{53.07}\% & \textbf{9.67}\% & \textbf{9.43}\%\\
    TempoBERT & 52.71\% & 51.28\% & 9.24\% & 8.85\%\\
    \bottomrule
\end{tabular}
\end{table}

\section{Conclusion}
\label{sec:conclusions}
In this paper, we presented a temporal contextual language model called TempoBERT which uses time as an additional context of texts. 
Our technique is based on modifying texts with time information and performing time masking---specific masking for the added time information. 
Using time masking, we model time as part of the context of texts. This enables us to both fill mask tokens while conditioning on time, and to predict the writing time of sentences. We experimented with two tasks: semantic change detection and sentence time prediction, and showed that both benefit from time masking. 
For semantic change detection, we proposed a method exploiting time masking that is especially strong for short-term corpora and outperformed existing state-of-the-art methods on diverse datasets, in terms of time, size, genre, and language. 
In addition, we experimented with small-sized TempoBERT models and showed their relatively strong performance on this task.
For sentence time prediction, time masking allows us to directly predict the time of sentences without additional fine-tuning. We showed strong performance compared to various baselines and studied different time masking settings to demonstrate the contribution of time masking to this task.


\section{Acknowledgements}
We thank Omer Levy and Yonatan Belinkov for their useful feedback.

\bibliographystyle{ACM-Reference-Format}
\balance 
\bibliography{main}

\end{document}